%% file: root.tex
\documentclass[letterpaper, 10 pt, conference]{ieeeconf}  % Comment this line out if you need a4paper

\IEEEoverridecommandlockouts                              % This command is only needed if 
\usepackage{times}
\usepackage{epsfig}
\usepackage{graphicx}
\usepackage{amsmath}
\usepackage{amssymb}

\usepackage{booktabs}
\usepackage{floatrow}
\usepackage{multirow}
\usepackage{adjustbox}
\newfloatcommand{capbtabbox}{table}[][\FBwidth]
\usepackage[dvipsnames]{xcolor}
\usepackage{blindtext}
\usepackage{cite}
\usepackage{siunitx}
\usepackage[format=plain,labelfont={bf,it},textfont=it,font=small,belowskip=-1.5pt]{caption}
\makeatletter
\let\NAT@parse\undefined
\makeatother
\usepackage{hyperref}
\hypersetup{
    colorlinks=true,
    linkcolor=blue,
    filecolor=magenta,      
    urlcolor=cyan,
    pdftitle={SeekNet},
    pdfpagemode=FullScreen,
    }
    
\newcommand{\sota}{state-of-the-art}
\newcommand{\algoname}{\textit{SeekNet}}
\newcommand\numberthis{\addtocounter{equation}{1}\tag{\theequation}}
\title{\LARGE \bf
\algoname: Improved Human Instance Segmentation and Tracking via Reinforcement Learning Based Optimized Robot Relocation}
\author{Venkatraman Narayanan, Bala Murali Manoghar, Rama Prashanth RV, Phu Pham and Aniket Bera \\
{Department of Computer Science, Purdue University, USA}\\
}

\begin{document}

\maketitle
\thispagestyle{empty}
\pagestyle{empty}

%%%%%%%%%%%%%%%%%%%%%%%%%%%%%%%%%%%%%%%%%%%%%%%%%%%%%%%%%%%%%%%%%%%%%%%%%%%%%%%%
\begin{abstract}

Amodal recognition is the ability of the system to detect occluded objects. Most {\sota} Visual Recognition systems lack the ability to perform amodal recognition. Few studies have achieved amodal recognition through passive prediction or embodied recognition approaches. However, these approaches suffer from challenges in real-world applications, such as dynamic obstacles. We propose \textbf{\algoname}, an improved optimization method for amodal recognition through embodied visual recognition. Additionally, we implement {\algoname} for social robots, where there are multiple interactions with crowded pedestrians. We also demonstrate the benefits of our algorithm on occluded human detection and tracking over other baselines. Additionally, we set up a multi-robot environment with {\algoname} to identify and track visual disease markers for airborne disease in crowded areas. We conduct our experiments in a simulated indoor environment and show that our method enhances the overall accuracy of the amodal recognition task and achieves the largest improvement in detection accuracy over time in comparison to the baseline approaches.
\end{abstract}

%%%%%%%%%%%%%%%%%%%%%%%%%%%%%%%%%%%%%%%%%%%%%%%%%%%%%%%%%%%%%%%%%%%%%%%%%%%%%%%%

\input{1-Intro}

\input{2-Related}
\input{4-Method}

\input{5-Segmentation}

\input{6-Navigation}
\input{7-Results}
\input{8-Conclusion}

{\small
\bibliographystyle{IEEEtran}
\bibliography{refs}
}

\end{document}

%% file: 1-Intro.tex
\section{Introduction}
\label{sec:intro}

Recent technologies in the field of robotics and AI have made remarkable advancements in the field of autonomous driving, mobile robots, social robots, etc. Most systems rely on a robust visual recognition system. Many recent works have improved Visual Recognition tasks such as Object recognition~\cite{qiao2020detectors, wang2020cspnet, tan2020efficientdet}, Semantic Segmentation~\cite{tao2020hierarchical, mohan2020efficientps, cheng2020panoptic}. Very few efforts have focused on amodal object recognition~\cite{yang2019embodied} and segmentation~\cite{qi2019amodal, follmann2019learning, zhang2019learning}. Amodal Visual Recognition is the ability of the system to perceive occluded objects~\cite{palmer1999vision}.

Some attempts were made to solve amodal recognition tasks by modeling it as an embodied recognition problem~\cite{yang2019embodied, das2018embodied}. These methods utilize the locomotive ability of a mobile robot to solve amodal recognition rather than passively attempting to predict the occluded object. Such a system works specifically well for social robots since most of the environment is occluded from the robots' field of vision (FOV).

The works~\cite{yang2019embodied, das2018embodied} have enabled a method to overcome the shortcomings of passively detecting occluded objects. The algorithms suffer from the following challenges:
\begin{itemize}
    \item They only work on a single target object at a time and expect only one instance of the target object within the searchable area.
    \item They lack the ability to track dynamic objects.
\end{itemize}

\begin{figure}[t!]
    \centering
    \includegraphics[width=\linewidth]{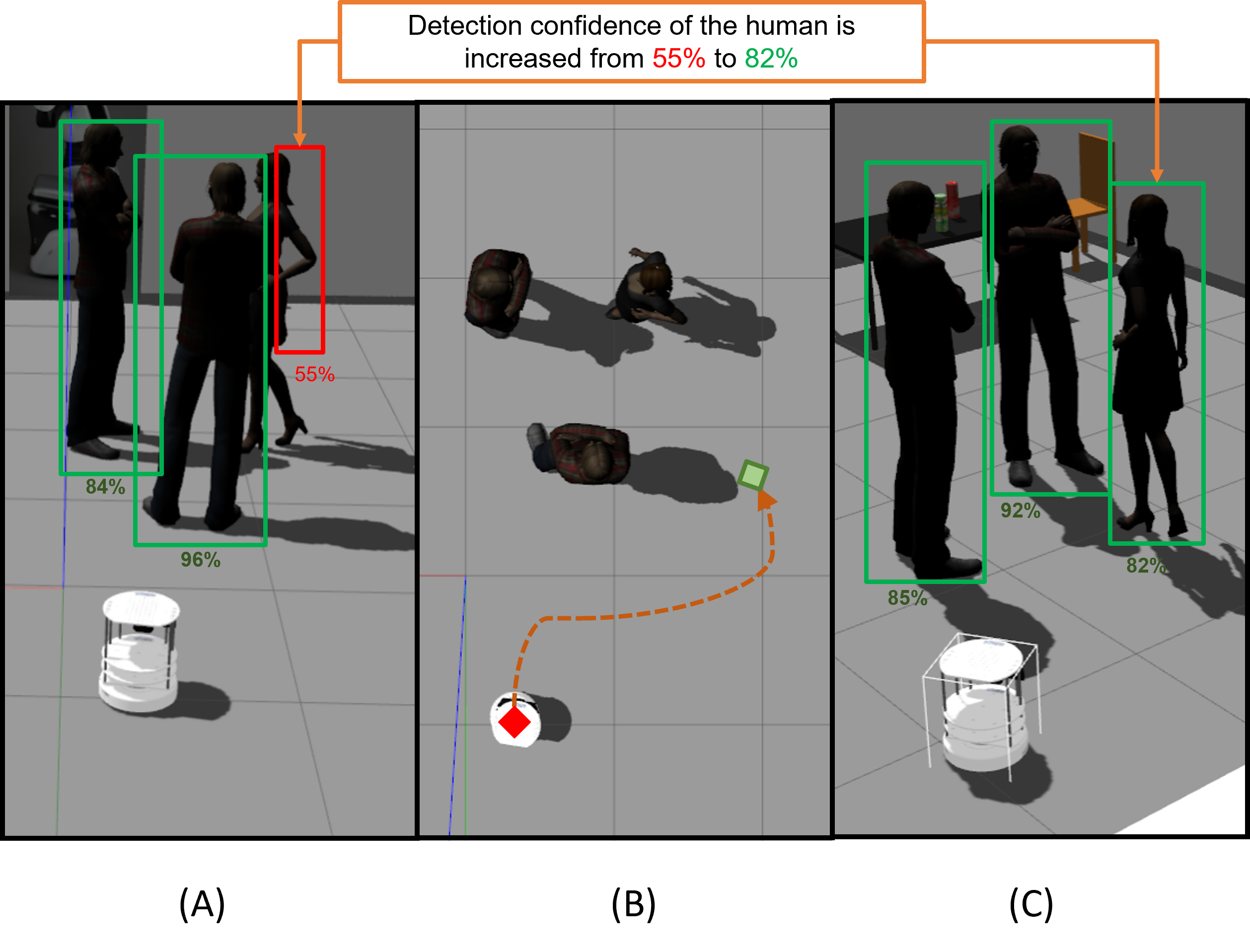}
        \caption{\textbf{\algoname} - an algorithm that achieves better Amodal Human Detection (A) through Embodied Recognition. (B) showcases our \textbf{\algoname} in action to move to a visually advantageous position depicted in (C), thereby improving the detection confidences of all humans.
        \vspace{-0.5cm}}
    \label{fig:Emotion}
    \vspace{-0.1cm}
\end{figure}

Furthermore, another issue with embodied recognition tasks that arise in highly dynamic environments is inefficient navigation planning in a multi-robot swarm configuration. The complexity of planning the navigation strategy for a multi-robot system in a dynamic environment is well-documented by \cite{fan2018crowdmove}. Current solutions lack the ability to track multiple goals for a multi-robot system. 

We propose {\algoname} to overcome such shortcomings with an ability to track dynamic objects and achieve embodied recognition tasks as depicted in figure \ref{fig:Emotion}. Since object detection is a vast topic and targeted algorithms are required to address the many sub-groups (or classes) under object detection, we mainly focus on embodied recognition for social robots. We tested our algorithm on a social robot, and since social robots primarily interact with dynamic humans in the environment, we designed an embodied recognition system that targets humans in the environment. To this end, our \textbf{main contributions} are:
\begin{itemize}
    \item We present a novel approach to perform the amodal segmentation of humans in a crowded environment and track them further.
    \item We provide an improved multi-robot navigation system based on policy networks that can explore a predefined environment to track the humans in it.
    \item Our approach can be used for Environmental Marker Monitoring for airborne diseases. We demonstrate an application with our {\algoname} to improve the pre-screening algorithms aimed at airborne disease detection, though our approach can be used in areas like hospitals, public places, etc.
\end{itemize}

The paper is organized as follows: Section 2 presents related work, section 3 gives an overview of our pipeline, section 4 describes each stage of our pipeline in detail, and finally, in section 5, we evaluate the theoretical and practical results of our work.

%% file: 2-Related.tex
\section{Related Work}
\label{Sec:Related_Work}

\subsection{Instance Segmentation}
\label{Sec:seg}

From a social robot navigation perspective, we have to model humans/pedestrians and other objects differently, and thus instance segmentation becomes an integral part of our pipeline. Broadly speaking, there are two broad approaches for instance, segmentation. The first approach uses a multi-stage pipeline by generating a pixel map of separate objects using the output from the object detection task \cite{he2017mask, dai2015convolutional, Girshick_Iandola_Darrell_Malik_2015, hariharan2014simultaneous, hariharan2015hypercolumns}. 

Another approach is to use a single network to collect high-level object information and low-level per-pixel information. The network results are combined to form a pixel map of individual objects. The features generated by this network are post-processed to obtain both object-level and pixel-level information \cite{dai2016instance, li2017fully, liu2017sgn, o2015learning, chennupati2021learning}.

Though there have been many advancements in segmentation, especially with the use of transformer networks \cite{wei2022contrastive, wang2022image, li2022mask}, which are not being considered here due to the computation complexity of transformer networks over convolutional networks and the specificity of heavily occluded Human segmentation compared to object segmentation. Furthermore, usage of vision transformer-based segmentation necessitates a large pool of training data \cite{li2022mask}. Hence, we use  Box2Pix~\cite{uhrig2018box2pix} approach, which presents a balanced fusion of object and pixel knowledge and produces accurate instance segmentation with an efficient single FCN forward pass and a single image pass post-processing. 

\subsection{Human Detection}
\label{Sec:seg}

For social navigation, the problem of occlusion is more pronounced because of frequent human-human interaction, even in a sparse crowd~\cite{socnav2,dorbala2020can}. For pedestrians, detection in such complex scenarios is usually achieved by using pose estimation techniques~\cite{bhattacharya2019take, bhattacharya2020step}. These networks are two-stage networks where the first stage extracts the skeleton information. The second stage combines pixel classification and poses information to generate a pixel map of individual humans. The approaches by ~\cite{zhang2019pose2seg,tripathi2017pose2instance} generate accurate masks even with heavy occlusion. Though the results are promising, these networks are computationally intensive, and the execution time exponentially increases with the number of humans in the scene. Thus these methods are unsuitable for integrating with a navigation scheme where real-time execution is necessary. To this end, our algorithm, \algoname uses a computationally efficient instance segmentation approach and leverages the movement capability of robots to achieve occlusion-free masks.

\subsection{Embodied Segmentation}
\label{Sec:emseg}

To better understand the shape of an object in case of occlusion, Qi et al.~\cite{qi2019amodal} trained a model to estimate the hidden region. Though they produce good results when the object's shape is complex, they are far from human-level performance. In order to accurately determine the shape of an object, Yang et al.~\cite{yang2019embodied} imitate the human ability to move and control the view angle actively. They introduced the task of Embodied amodal segmentation and addressed the problem using Embodied Mask-RCNN. This approach is trained for static objects, but moving humans are the prime targets in the case of social navigation. To this end, our approach \algoname is trained for such dynamic environments and can maintain a constant distance from the target.

% \subsection{AI models for COVID Prognosis and Diagnosis}
% \label{Sec:Rel_covid}

% There are numerous AI models available to classify an infectious person and a non-infectious person. The improvement in the accuracy of these models is also fast-paced. ~\cite{santosh2020ai, esteva2019guide, ulhaq2020computer} explore different AI approaches related to COVID-19 detection. However, these models do not scale well, and there is a high risk of bias that raises concerns about being used in daily practice. To mitigate this,~\cite{Somboonkaew:20} uses multiple sensors to improve accuracy. Various methods to fuse different algorithms have also been explored in ~\cite{8857027, negishi2020contactless}. These methods use late fusion due to the unavailability or sparse availability of multi-modal data, such as temperature,  heart rate, respiratory rate,  cough signature, etc.,  for the same human subject. Uniformly all these models need a specific environment setting to get the best accuracy and are far from deployment in public places. Our model \algoname could be leveraged for deploying these contactless diagnosis models in public places as it can seclude a single person in a group and keep him in focus for the entire course of measurement and access the state of health of a person.

\subsection{Social Robots and Embodied Navigation}
Embodied Navigation is approached with the goal of exploring new (indoor) environments in the shortest time to identify specific targets of interest within the environment. Existing architectures utilize reinforcement learning for exploration and navigation to maximize coverage \cite{ye2021auxiliary, ye2020auxiliary, ramakrishnan2021exploration, chen2019learning, ramakrishnan2020occupancy}. The main disadvantage of these systems is that they assume the environment to be static.
Indoor environments, often, robots need to coexist with humans and are expected to follow socially acceptable navigation and interaction. Several works try to adapt navigation in crowded scenarios by considering social norms and human emotions \cite{vega2019socially, narayanan2020proxemo, vega2021towards, narayanan2020ewarenet}. Affect recognition from features such as facial expressions, gestures, and walks has been addressed in the literature surveyed in ~\cite{bhattacharya2020generating, banerjee2020learning,  randhavane2019modeling, bera2020you}. Multi-modal and context-aware affect recognition models are also available,~\cite{m3er,multimodal1,emoticon, mittal2020emotions}, that incorporate such emotions in the social navigation pipeline. Our work presents an approach to explore and target dynamic objects of interest (humans).

%PLEASE WRITE ONE LINE TO CONCLUDE THIS (WHAT MAKES THESE WORK DIFFERENT FROM YOURS?)

%% file: 4-Method.tex
\section{Overview and Methodology}
\label{sec:methodology}
\begin{figure*}
\centering
\includegraphics[width=1\textwidth]{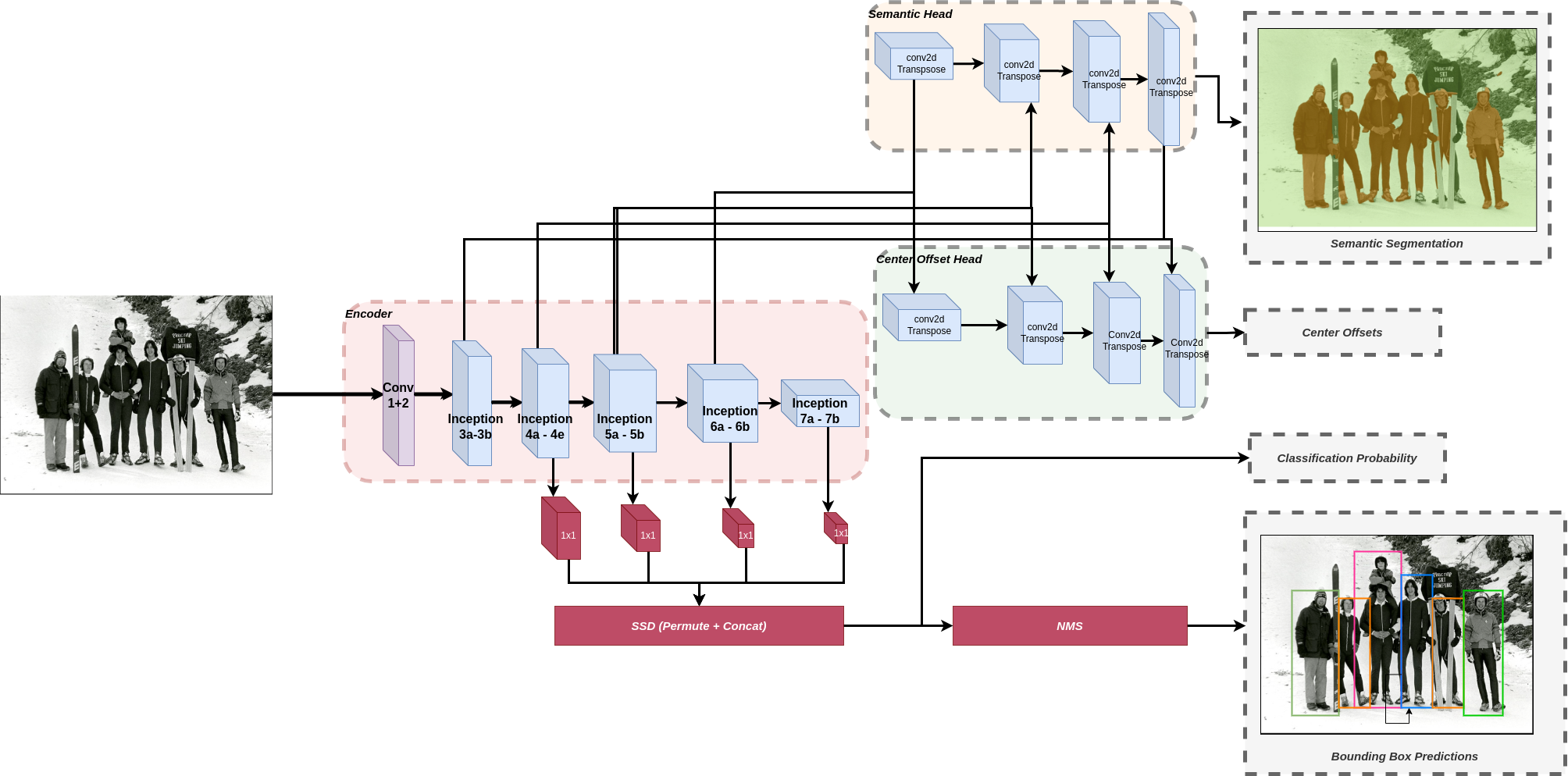}
\caption{\textbf{Segmentation}: Our proposed pipeline for amodal human detection and tracking. We build on~\cite{uhrig2018box2pix} and adapt their network architecture to generate three types of outputs: semantic segmentation, center offsets, object bounding box, and object classification confidence}
\label{fig:bd}
\end{figure*}

The primary goal for {\algoname} is to learn an optimal solution that improves the weakly learned detectors. Specifically, our navigation pipeline aims at improving Visual Recognition algorithms in an Embodied Recognition setup. Our {\algoname} relies on commodity RGB cameras onboard a mobile robot with other components necessary for robot perception and navigation. Our Visual Perception system consists of Amodal Recognition and Amodal Segmentation to identify potentially occluded objects, refined by our novel navigation system to improve the detection confidence (or accuracy) by maneuvering the robot to a more advantageous position. Our system relies on a mobile robot's ability to reposition itself to complete the Visual Recognition task at hand better. Figure~\ref{fig:bd} provides a brief overview of our {\algoname} system.
%% These perceived emotions are then used to compute variable proxemic constraints in order to perform socially aware navigation through a pedestrian environment.\\
%% Figure \ref{fig:overview} illustrates how we incorporate {\algoname} into an end-to-end \textit{emotionally-guided} navigation pipeline.

The following subsections will describe our approach in detail. We discuss the details of the datasets used to train our perception and policy network, along with other processing techniques (if any) used. We also provide details on our Amodal Recognition and Segmentation routine, where we also briefly discuss our human detection and segmentation routine from an RGB camera. Finally, we discuss our navigation system.

%% file: 5-Segmentation.tex
\subsection{Segmentation}
\label{sec:seg}

\subsubsection{Network Structure}
\label{sec:network}

Since our objective is to deploy \algoname on a robot, the model must achieve sufficient frame rates for use in the navigation pipeline. We build on~\cite{uhrig2018box2pix} and adapt their network architecture to generate three types of outputs: instance segmentation, object classification, and object classification confidence. Figure \ref{fig:bd} provides an overview of our segmentation pipeline.

We retain the~\cite{uhrig2018box2pix} modifications from the GoogLeNet’s inception module. The additional inception modules added to the backbone achieve a larger receptive field that helps identify humans/pedestrians close to the robot. To predict the actual box parameters and box classes, we add $1 \times 1$ convolutions from different levels of backbone layers and recursively compute the receptive field theoretically, similar to \cite{uhrig2018box2pix}.
\vspace{-3mm}
\begin{align}
\label{eq:rf}
RF_{out} = (RF_{in} - 1) \times s + k
\end{align}

where $RF$ is the input and output receptive field, $s$ is the stride of the corresponding layer, and $k$ is the kernel size. We maintain the theoretical receptive fields to be twice that of the maximum value of the height or width of prior boxes while assigning it to a specific layer. This is done to compensate for the reduction in the receptive field during training, as suggested by~\cite{luo2016understanding}.

% \begin{figure}[h!]
% \centering
% \includegraphics[width=1.0\textwidth]{images/relative_change_metric.png}
% \caption{Relative Change Metric}
% \label{fig:rcm}
% \end{figure}

The semantic class and center offset class is predicted using skip connections from inception modules of corresponding layers. It consists of $1 \times 1$ convolutions with element-wise addition and deconvolutions to upscale the low-resolution feature maps sequentially. 

\subsubsection{Loss Formulation}

We use a hybrid loss (equation \ref{eq:loss}) to train our network. The hybrid loss is a weighted combination of losses for each of the sub-tasks (semantic, offsets, bounding box, classification) performed by our network using the approach presented by Kendall et al. \cite{kendall2018multi} to learn task uncertainties $\sigma$

\begin{equation}
    \label{eq:loss}
    \begin{split}
        L_{total} &= \frac{1}{\sigma^{2}_{sem}} \cdot L_{sem} + \log \sigma_{sem} \\
        &+ \frac{1}{\sigma^{2}_{off}} \cdot L_{off} + \log \sigma_{off} \\
        &+ \frac{1}{\sigma^{2}_{bbox}} \cdot L_{bbox} +  \log \sigma_{bbox} \\
        &+ \frac{1}{\sigma^{2}_{cls}} \cdot L_{cls} +  \log \sigma_{cls}\\ 
    \end{split}
\end{equation}

We use a standard cross-entropy loss for semantic segmentation, $L_{sem}$. L2 regression loss for center offset vectors $L_{offsets}$. Both semantic segmentation loss and center offset loss are normalized over the number of valid pixels. $L_{bbox}$ is the L2 regression loss for bounding box parameters ($x_{min}$, $y_{min}$, $x_{max}$, $y_{max}$). For classification $L_{cls}$, we use Focal Loss \cite{lin2017focal} to counter the imbalance between foreground and background classes.
\\
\subsubsection{SSD Adaptation and Instance Segmentation}
% CAN YOU INTRODUCE THE OCHUMAN DATASET AND WHY IT'S RELEVANT HERE. I DON'T THINK THE PAPER HAS TALKED ABOUT IT BEFORE
In heavily crowded scenarios, humans of often occluded and the bounding boxes are relatively small and incomplete. In such cases, the intersection-over-union (IoU) for small, incomplete prior boxes results in bad object detection performance. So to mitigate the problem of low coverage, we use the relative box parameter instead of IoU. This also helps in the training process since we match the loss (based on corner offsets) and generation metric in a common space. The relative change between  $b_{prior}$ of size ($x_{min}$, $y_{min}$, $x_{max}$, $y_{max}$) and annotated ground truth box $b_{GT}$ is given by equation \ref{eq:relch}

\begin{align*}
d_{change} &= \sqrt{ \frac{\Delta y_{tl}^{2}}{h_{GT}} + \frac{\Delta x_{tl}^{2}}{w_{GT}} + \frac{\Delta y_{br}^{2}}{h_{GT}} + \frac{\Delta x_{br}^{2}}{w_{GT}}} \numberthis \label{eq:relch}
\\
where&, \notag \\
w_{GT} &= x_{max} - x_{min} \notag \\
h_{GT} &= y_{max} - y_{min} \notag \\
\end{align*}

$\Delta x$ and $\Delta y$ are the absolute difference in the two boxes' $x$ and $y$ parameters, and $tl$, $br$ represent top-left and bottom-right positions.

In order to densely cover both small and large objects, we use $21$ prior anchor boxes. The number and dimensions of prior bounding boxes are found by running k-means clustering on the training bounding boxes. The details of this method are described in ~\cite{redmon2017yolo9000}. It is worth mentioning that using $21$ bounding boxes for our dataset works better than 5 as recommended in ~\cite{redmon2017yolo9000}.

We combine the output from three outputs: semantic class, center offset vectors, and object detection with bounding boxes to generate instance segmentation output as proposed in~\cite{uhrig2018box2pix}.

%% file: 6-Navigation.tex
\subsection{Embodied Recognition}
\label{sec:nav}
We train a policy network to achieve the embodied recognition task. Based on the human detection confidence of our amodal recognition  (section \ref{sec:network}), we identify potential goal points to pursue and refine using our embodied recognition system. Our identification process is based on weak detection confidence below a threshold $(\lambda)$. We build our policy network upon ~\cite{yang2019embodied,fan2018crowdmove}. Our policy network receives the LiDAR scans and each human segmentation masks from the amodal recognition system and outputs probabilities
over the action space considered for the navigation task. 

\textbf{Action Space:} The action space is a set of permissible robot velocities in continuous space. The action velocities consist of translational and rotational velocities. We set the bounds on translational velocity, $v \in [0.0, 1.0]$ and rotational velocity, $w \in [-1.0, 1.0]$ to accommodate the robot kinematics. We sample actions at step $t$ using the equation \ref{eq:act}

\begin{align}
\label{eq:act}
a_t = \pi(l_0, l_1, l_2, h_0, h_1, ..., h_i)
\end{align}
where $l_0, l_1, l_2$ represent the three consecutive processed LiDAR scan frames and $h_0, h_1, ..h_t$ represent the historical and human segmentation masks concatenated together.

\begin{figure*}
\centering
\includegraphics[width=1\textwidth]{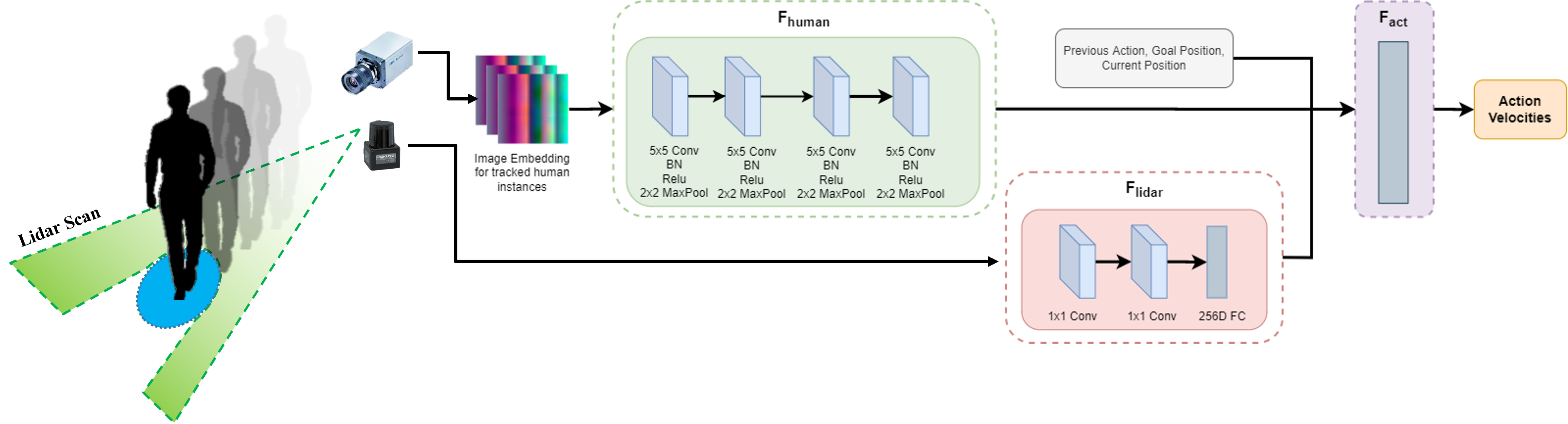}
\caption{\textbf{Embodied Navigation}: Our Embodied Navigation pipeline takes in the tracked human instances, LiDAR scans, and Robot's positional information as input to a policy network. The policy network outputs the \textit{robot velocity vectors} as action space.}
\label{fig:RLNet}
\end{figure*}

\textbf{Policy Network}: The policy network has three components $\{f_{human}, f_{lidar}, f_{act}\}$ (showcased in figure \ref{fig:RLNet}). $f_{human}$ represents the network for encoding the human segmentation masks. $f_{lidar}$ encodes the LiDAR frames and $f_{act}$ represents the network that outputs action velocities for the robot based on encoded segmentation masks, encoded lidar frames, along with previous position and velocities of the robot.

We resize the masks from our Instance Segmentation network to $244 \times 244$ to adapt to the architecture of the human segmentation mask encoding component ($f_{human}$) of our policy network. We pass them to $f_{human}$, which consists of four $5\times5$ Conv, BatchNorm, ReLU. Each Conv block is followed by a $2\times2$ MaxPool blocks, producing an encoded human segmentation mask $ z^{img}_{t} = f_{human}([h_0, h_1, h_i])$

We process the three consecutive lidar frames by passing them through two $1\times1$ Conv, followed by a $256D$ fully-connected (FC) layer. The lidar frames are encoded as $z^{lidar}_{t} = f_{enc}([l_0, l_1, l_2])$. The $f_{act}$ is a multi-layer perceptron (MLP) network, with 1 $128D$ FC hidden layer, and finally produces the action velocities. $f_{act}$ takes in the encoded human trajectories, $z^{img}_{t}$, lidar encodings, $z^{lidar}_{t}$, previous velocity, $v_{t-1}$, goal position, $s_g$, and current robot position, $s_t$,  to predict the robot velocities at time $t$, given by equation \ref{eq:velocity}.

\begin{align}
\label{eq:velocity}
v_t = f_{act}([z^{img}_{t}, z^{lidar}_{t}, v_{t-1}, s_g, s_t])
\end{align}

$v_t$ is then sent to a linear layer with softmax to derive the probability distribution over the action space from which the action is sampled. We learn $\{f_{human}, f_{lidar}, f_{act}\}$ via reinforcement learning. 

\textbf{Rewards}: Our reward function for the policy network is inspired from \cite{fan2018crowdmove}. We aim to arrive at an optimal strategy to avoid collisions during navigation while ensuring that we improve the targeted object's detection confidence (human). The reward function to achieve the mentioned goals is given in equation \ref{eqn:overallreward}

\begin{equation}
    \label{eqn:overallreward}
    r^t = r_c^t + r_w^t + r_h^t
\end{equation}
The reward $r$ at time $t$ is a combination of the reward for avoiding collisions, $r_c$, a reward for smooth movement, $r_w$, and a reward for improving the detection confidence, $r_h$. 

The penalty for colliding with obstacles is given by equation \ref{eqn:penalrew}.

\begin{equation}
    \label{eqn:penalrew}
    r_c^t = \begin{cases}
            r_{collision}, & \text{if robot collides}.\\
            0, & \text{Otherwise}.
            \end{cases}
\end{equation}
For ensuring smooth navigation, the penalty for large rotational velocities is given by equation \ref{eqn:smoothrew}.

\begin{equation}
    \label{eqn:smoothrew}
    r_w^t = \begin{cases}
            w_w|w^t|, & \text{if $|w^t| > 0.7$}.\\
            0, & \text{Otherwise}.
            \end{cases}
\end{equation}
To ensure that we progressively reposition the robot to improve detection confidence on the targeted object, we reward the system based on \ref{eqn:emorew}. The penalty is only applied when the robot is actively pursuing a target.

\begin{equation}
    \label{eqn:emorew}
    r_e^t = \begin{cases}
            r_p, & \text{if $p_{t}^{h^i} > p_{t-1}^{h^i}$}.\\
            r_n, & \text{Otherwise}.
            \end{cases}
\end{equation}

In our implementation, we use $r_{arrival} = 15, w_g = 2.5, r_{collision} = -15, w_w = -0.1, r_p = 2.5, r_n = -0.5, \xi = 0.1$

%% file: 7-Results.tex
\section{Experiment and Results}
\label{Sec:Results}
\subsection{Metrics}
\label{Sec:metrics}
We evaluate our amodal recognition efficiency based on classification accuracy $(Acc_{cls})$, and segmentation accuracy as mean Intersection-over-Union (IoU) on the first frame of detection. We also report the tracking accuracy $(Acc_{tr})$ to evaluate our amodal recognition system. We evaluate the embodied recognition system in terms of \textit{change in classification accuracy} $(\Delta_{acc}^{h})$.

\subsection{Datasets}\label{sec:data}

In this work, we use datasets specially designed for detecting humans with heavy occlusions.

\textbf{OCHuman}: OCHuman \cite{zhang2019pose2seg} is a large dataset designed for all three tasks: detection, pose estimation, and instance segmentation. This dataset captures severe occlusion between human bodies is often encountered in real life. It contains $8110$ detailed annotated human instances within $4731$ images. The dataset primarily emphasizes occlusions to encourage the development of algorithms more suited for practical and real-life situations.

\textbf{JTA Dataset}: JTA (Joint Track Auto) \cite{fabbri2018learning} dataset is a massive collection of pedestrian pose estimation and tracking in urban scenarios. The data is created by exploiting the highly photorealistic video game Grand Theft Auto V developed by Rockstar North. The dataset contains $512$ video clips from several scenarios in urban environments. The dataset covers variations in illumination and a variety of view angles. It also covers indoor and outdoor scenarios with natural actions like sitting, running, chatting, etc., in a typical crowded environment. The clips are precisely annotated with values of visible and occluded body parts, people tracking with 2D and 3D coordinates.

\subsection{Implementation Details}
\label{sec:impdet}
\textbf{Amodal Recognition}: We train our pipeline on dataset described in section (\ref{sec:data}) with a train-validation split of 90\%-10\%. 
We use ADAM \cite{kingma2014adam} optimizer, with decay parameters of ($\beta_{1} = 0.9 $ and $\beta_{2}= 0.999$) to train our networks. We set the initial learning rate as $0.009$ with $10\%$ decay every 250 epochs. The models were trained with the hybrid loss detailed in section \ref{sec:nav}.
% The training was run for $2000$ epochs, or when training and evaluation loss does not change for more than $5$ epochs.
We used two NVIDIA RTX 2080 Ti GPUs having 11GB of GPU memory each and 64 GB of RAM to perform our experiments.

\textbf{Embodied Recognition}: We train our embodied recognition policy network on a simulation environment generated using the Stage Mobile Robot Simulator \cite{vaughan2008massively}. We generate multiple scenarios (see in figure \ref{fig:tain}) with obstacles to train our policy network. We use RMSProp \cite{hinton2012neural} for training our policy network with learning rate $0.00004$ and $\epsilon = 0.00005$.

\begin{table}[tb]
\centering
\vspace{0.25cm}
\resizebox{\columnwidth}{!}{
\begin{tabular}{@{}cc|c|c|c|ccc@{}}
\toprule
\multicolumn{2}{c}{Moving Path}  & \multicolumn{3}{c}{Amodal Recognition} &  \multicolumn{3}{c}{Embodied Recognition} \\ \hline \hline
Training & Testing  & $Acc_{cls}$ & mIoU & $Acc_{tr}$ & \multicolumn{3}{c}{$\Delta_{acc}^{h}$ after ms}\\
\multicolumn{2}{l|}{}  &  &  &  & 80 & 160 & 320  \\
\hline \hline
Passive & Passive & 87.5 & 72.4 & 71.5 & - & - & -\\ \hline
ShortestPath & Passive & 87.4 & 72.5 & 71.6 & 87.6 & 88.3 & 90.1\\ \hline
ShortestPath & RandomPath & 87.5 & \textbf{72.6} & 71.5 & 87.8 & 88.2 & 89.1\\ \hline
ShortestPath & ShortestPath & \textbf{87.7} & 72.5 & 71.4 & 87.9 & 88.1 & 89.8\\ \hline
ShortestPath & {\algoname} (ours) & 87.6 & \textbf{72.6} & 71.6 & 88.2 & 89.1 & 90.3\\ \hline
{\algoname} (ours) & {\algoname} (ours) & 87.6 & 72.5 & \textbf{72.6} & 88.3 & \textbf{89.7} & \textbf{90.5} \\
\hline \hline

\bottomrule
\end{tabular}
}
\caption{Comparison of {\algoname} with various baselines using scenarios $5$ and $6$ from figure \ref{fig:tain}. We report the metrics mentioned in section \ref{Sec:metrics} for all the baselines described \cite{yang2019embodied}. We can see that our {\algoname} has the \textbf{best \textit{change in classification accuracy}} ($\Delta_{acc}^h$) across all our experiments.}
\label{tab:amodalres}
%\vspace{-0.75cm}
\end{table}

\subsection{Results and analysis}
\begin{figure*}
\centering
\includegraphics[width=1\textwidth]{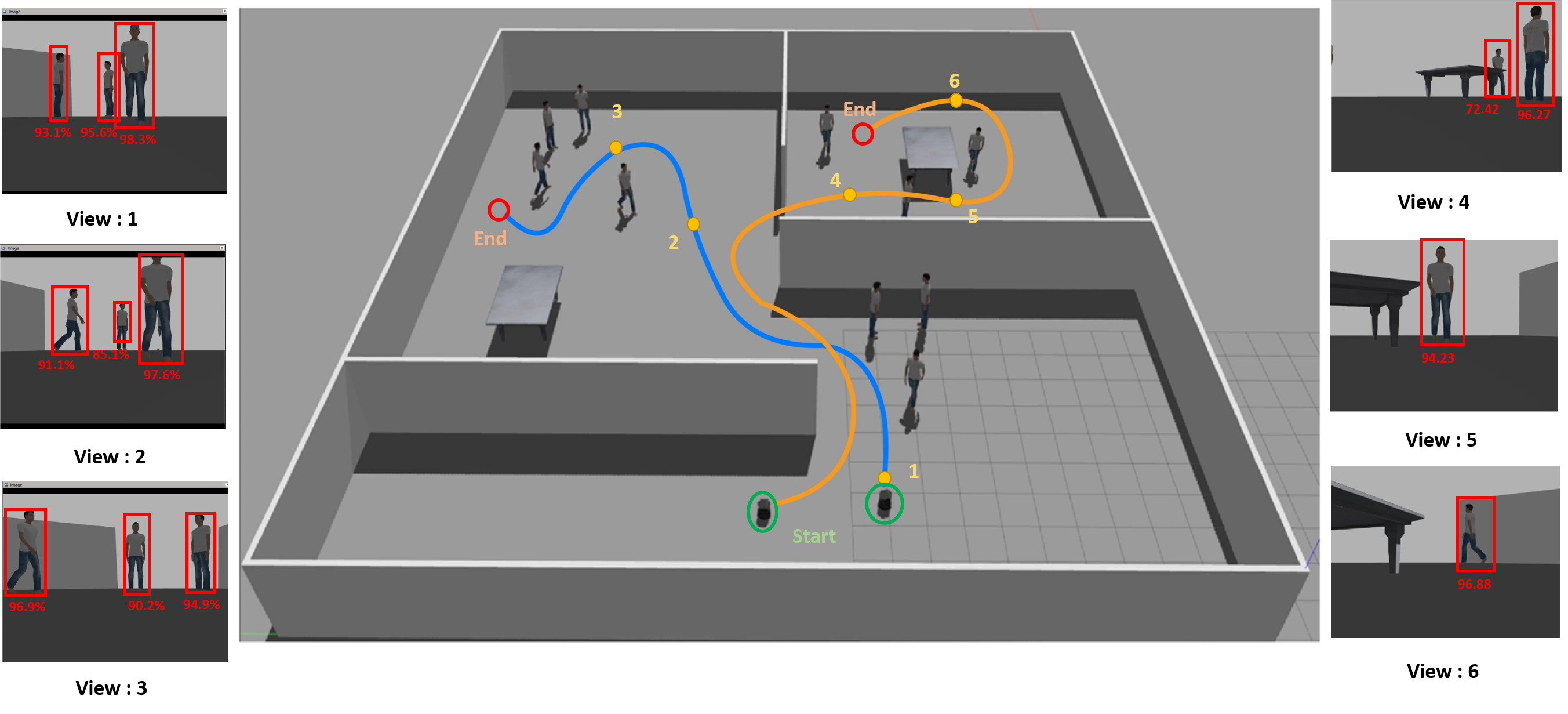}
\caption{\textbf{Embodied Navigation Result}: In the above figure, the \textcolor{blue}{blue} and \textcolor{orange}{orange} lines represent the trajectories followed by the robots, namely robot 1 and robot 2 respectively. For simplicity, let's consider robot 1. The path taken by the robot is not the shortest path to the goal position. Instead, the robot takes a path that maximizes human detection confidence. The viewpoints on the path are marked with \textcolor{yellow}{yellow} dots, and the corresponding detection confidence is displayed from the robot's perspective. In view-1, all humans are detected with good confidence. In view-2, the human at the center is identified with lower confidence when the robot is moving towards its goal. Thus, the robot moves near the humans to view-3, where the human detection confidence is improved.}
\label{fig:result}
\end{figure*}

\begin{figure}[t!]
\centering
\includegraphics[width=1\textwidth]{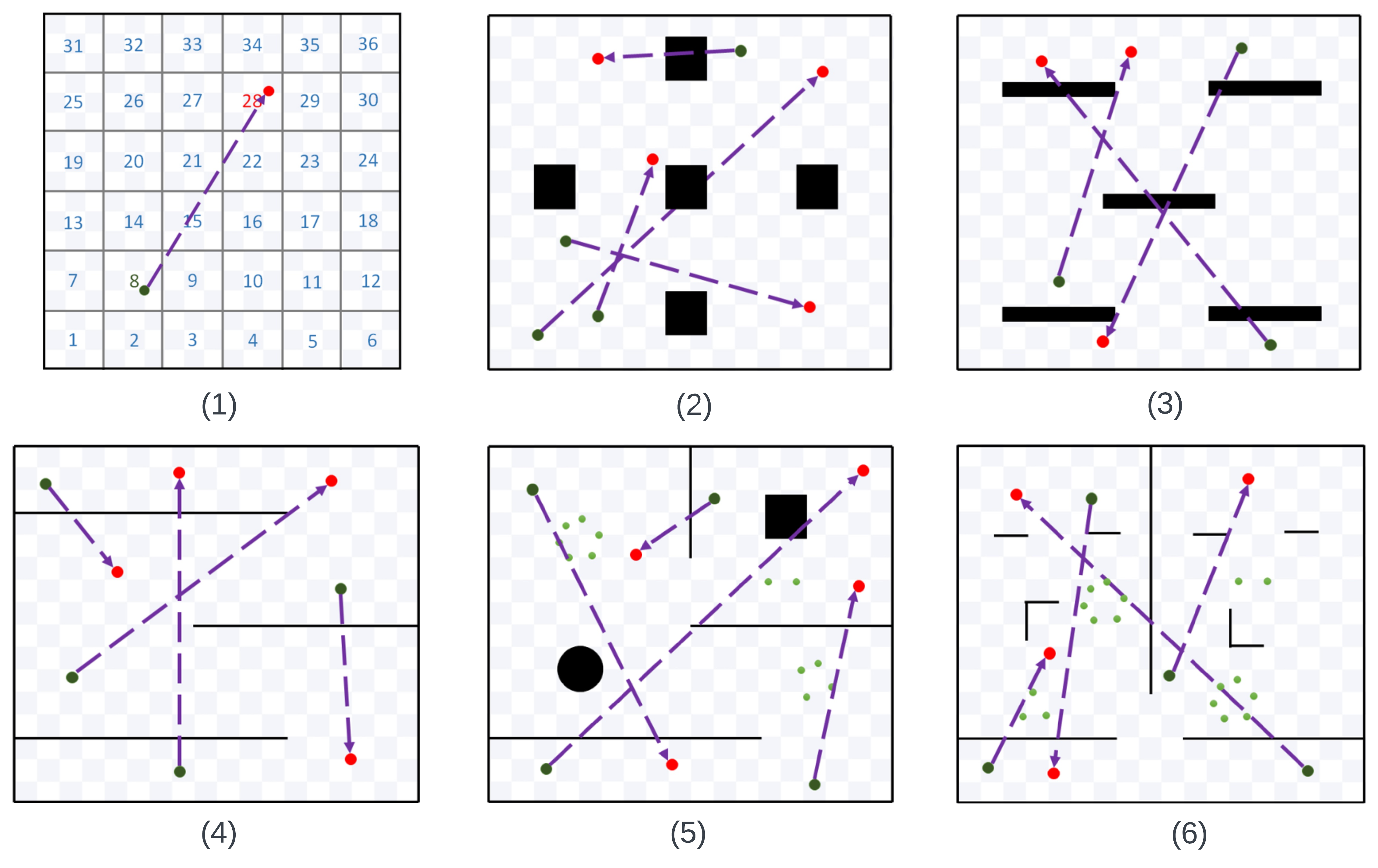}
\caption{The various scenarios used to train our Embodied Navigation pipeline. The shapes (square, rectangle, round) filled in black represent the geometries of the obstacles. The black, solid lines are the walls. The light green dots are the humans standing in the scene. The green and red dots indicate the start and goal positions of the agent. We model the world as a grid world shown in figure 1. As mentioned in section \ref{sec:impdet}, several configurations of parameters within a scenario are tweaked to generate multiple training/testing scenarios based on figures (1)-(6).}
\label{fig:tain}
\end{figure}

We illustrate the quantitative results from our experiments in Table \ref{tab:amodalres}. We use various scenarios from figure \ref{fig:tain} to perform our comparison studies. We compared our approach to several baseline methods using different training and testing paths. The baseline methods include: 
\begin{itemize}
\item Passive/Passive: the agent does not move during both training and testing.

\item ShortestPath/Passive: the agent moves along the shortest path during training but does not move during testing

\item ShortestPath/RandomPath: the agent moves randomly during testing to determine whether strategic moves are required for embodied amodal recognition. 
\item  ShortestPath/ShortestPath: the agent moves along the shortest path during both training and testing
\end{itemize}
For a complete view of these baselines, please refer to the detailed explanations in \cite{yang2019embodied}.

We report the metrics mentioned in section \ref{Sec:metrics} for all the above-mentioned baseline models. The experiment results demonstrate that {\algoname} has the best \textit{change in classification accuracy} across all our experiments. For amodal recognition task,  {\algoname} does not achieve the best accuracy, but its performance is on par with the "oracle-like" baseline ShortestPath/ShortestPath. It is called "oracle-like" since it must know the entire structure of the 3D environment in order to compute the shortest path \cite{yang2019embodied}. In contrast, {\algoname} does not have complete knowledge of the environment, but the model accumulates it over time. The increase in classification accuracy demonstrates that {\algoname} eventually learns the environment and selects the best actions. The visual results are demonstrated in figure \ref{fig:result}.

\begin{table}[h]
\centering
\vspace{0.25cm}
\resizebox{\columnwidth}{!}{
\begin{tabular}{@{}|c|c|c|c|@{}}
\toprule
Metric  & Passive/Passive & {\algoname}/{\algoname} (ours)  \\ \hline \hline
$Acc_{cls}$ & 87.4 & \textbf{87.9} \\ \hline
$mIoU$ & 72.5 & \textbf{72.9} \\ \hline
$Acc_{tr}$ & 71.6 & \textbf{72.7} \\ \hline
$Acc_{C19}$ & 67.2 & \textbf{71.3} \\ \hline
$\Delta_{acc}^{h}$ after 80 ms & - & 88.1 \\ \hline
$\Delta_{acc}^{h}$ after 160 ms & - & 88.5 \\ \hline
$\Delta_{acc}^{h}$ after 320 ms & - & 89.1 \\ \hline
\bottomrule
\end{tabular}
}
\caption{Our experimental results from COVID-19 detection system. We perform the experiment similar to our Embodied Human Detection \& Segmentation routine explained above, with the exception that we focus on the COVID-19 symptom detection confidence. We use an ensemble of algorithms that screens for COVID-19 symptoms from a multitude of sensors. We see \textit{\textbf{significant boost COVID-19 screening confidence when using our algorithm}}.}
\label{tab:covidres}
%\vspace{-0.75cm}
\end{table}

\subsection{Using \algoname~for Environmental Marker Monitoring}
As mentioned earlier, we experiment {\algoname} for tracking and monitoring humans for visual markers of air-borne diseases in indoor environments. We perform the experiment similar to our Embodied Human Detection \& Segmentation routine explained above, with the exception that we focus on individuals with markers of air-borne diseases instead of just human detection. We use an ensemble of off-the-shelf algorithms that screens for markers of infectious disease such as flu, cough, cold, and fever. The results are reported in table \ref{tab:covidres}. We see significant potential in using \algoname~to passively screen for such visual markers to prevent the spread of such air-borne infections. It is important to note that we still use an amodal human detection pipeline to detect and target human symptom screening.

%%%%%%%%%%%%%%%%%%%%%%%%%%%%%%%%%%%%%%%%%%%%%%%%%%%%%%%%%%%%%%%%%%%%%%%%%%%%%%%

%% file: 8-Conclusion.tex
\section{Conclusion and limitations}
In this paper, we have proposed a novel approach called {\algoname} for embodied recognition tasks, specifically focused on amodal human detection in a crowded environment. {\algoname} can overcome the shortcomings of passively detecting occluded objects with an ability to track dynamic objects. Since object detection is a vast topic and targeted algorithms are required to address the many sub-groups (or classes) under object detection, we mainly focused on embodied recognition for social robots. We tested our algorithm on a simulated social robot, and since social robots primarily interact with dynamic humans in the environment, we designed an embodied recognition system that targets humans in the environment. 

\textbf{Limitations:} Currently, the robot greedily chooses a position from which the humans can be clearly perceived. A social scene is a complex scenario, and so the greedy approach does not reflect the optimal actions. Moreover, we tested our method in a simulated environment, and further development and validations need to be done to generalize and adapt our approach to real-world scenarios.

\section{Acknowledgements:} The research reported in this publication was supported by funding from the National Science Foundation EAGER COVID-19 Grant.